\documentclass[10pt,conference]{IEEEtran}

\usepackage{cite}
\usepackage{hyperref}
\usepackage[flushleft]{threeparttable} 

\ifCLASSINFOpdf
   \usepackage[pdftex]{graphicx}
   \graphicspath{{images/}}
   \DeclareGraphicsExtensions{.pdf,.jpeg,.png}
\else
   \usepackage[dvips]{graphicx}
   \graphicspath{{../images/}}
   \DeclareGraphicsExtensions{.eps}
\fi
\usepackage{booktabs}
\usepackage{makecell}
\usepackage[cmex10]{amsmath}
\usepackage{amsthm}

\usepackage{algorithmic}

\usepackage{array}

\ifCLASSOPTIONcompsoc
  \usepackage[caption=false,font=normalsize,labelfont=sf,textfont=sf]{subfig}
\else
  \usepackage[caption=false,font=footnotesize]{subfig}
\fi
\usepackage{url}

\newif\iffinal
\finaltrue
\newcommand{\cmtid}{52}

\iffinal
\else
\usepackage[switch]{lineno}
\fi

\begin{document}
%
\title{STN PLAD: A Dataset for Multi-Size Power Line Assets Detection in High-Resolution UAV Images}


\iffinal




%
\author{\IEEEauthorblockN{André Luiz Buarque Vieira-e-Silva\IEEEauthorrefmark{1},
Heitor de Castro Felix\IEEEauthorrefmark{1},
Thiago de Menezes Chaves\IEEEauthorrefmark{1}, \\
Francisco Paulo Magalhães Simões\IEEEauthorrefmark{1}\IEEEauthorrefmark{2},
Veronica Teichrieb\IEEEauthorrefmark{1},
Michel Mozinho dos Santos\IEEEauthorrefmark{3}, \\
Hemir da Cunha Santiago\IEEEauthorrefmark{3},
Virginia Adélia Cordeiro Sgotti\IEEEauthorrefmark{3},
and
Henrique Baptista Duffles Teixeira Lott Neto\IEEEauthorrefmark{4}}
\IEEEauthorblockA{\IEEEauthorrefmark{1}Voxar Labs, Centro de Informática,
Universidade Federal de Pernambuco,
Recife, Brazil\\ \{albvs,hcf2,tmc2,vt\}@cin.ufpe.br}
\IEEEauthorblockA{\IEEEauthorrefmark{2}Departamento de Computação, Universidade Federal Rural de Pernambuco, Recife, Brazil\\
francisco.simoes@ufrpe.br}
\IEEEauthorblockA{\IEEEauthorrefmark{3}In Forma Software, Recife, Brazil\\
\{mmozinho,hsantiago\}@informasoftware.com.br, vsgotti@informa.com.br}
\IEEEauthorblockA{\IEEEauthorrefmark{4}Sistema de Transmissão Nordeste - STN, Recife, Brazil\\
hlott@stnordeste.com.br}}

\else
  \author{Sibgrapi paper ID: \cmtid \\ }
  \linenumbers
\fi

\maketitle

\begin{abstract}
Many power line companies are using UAVs to perform their inspection processes instead of putting their workers at risk by making them climb high voltage power line towers, for instance. A crucial task for the inspection is to detect and classify assets in the power transmission lines. However, public data related to power line assets are scarce, preventing a faster evolution of this area. This work proposes the STN Power Line Assets Dataset, containing high-resolution and real-world images of multiple high-voltage power line components. It has 2,409 annotated objects divided into five classes: transmission tower, insulator, spacer, tower plate, and Stockbridge damper, which vary in size (resolution), orientation, illumination, angulation, and background. This work also presents an evaluation with popular deep object detection methods and MS-PAD, a new pipeline for detecting power line assets in hi-res UAV images. The latter outperforms the other methods achieving 89.2\% mAP, showing considerable room for improvement. The STN PLAD dataset is publicly available at \url{https://github.com/andreluizbvs/PLAD}.
\end{abstract}



\section{Introduction}

Nowadays, practically all human activities depend on the constant availability of electricity. Power transmission lines, which have an essential role in this task, are constantly exposed to the depreciating action of the environment. They have components that may break, rust, loosen or even go missing. The malfunction of such equipment affects the electricity grid, causing inefficiency in the power transmission and, sometimes, blackouts. According to \cite{nguyen2018automatic}, most of the power grids today are interconnected. Thus, these blackouts can initiate others, affecting even larger regions, like a cascade effect \cite{pradeep2012high}. That can trigger catastrophic consequences such as shutting down hospitals, production at water supplies companies, and telecommunication services \cite{castillo2014risk}, which leads to significant economic losses for the energy company and, ultimately, severe social impacts \cite{bruch2011power,li2020state}. According to Bruch et al. \cite{bruch2011power}, a power cut of only 30 minutes in the USA results in an average loss of over 15 thousand US dollars for midsize and large industrial clients and a loss of more than 90 thousand US dollars for an eight-hour interruption \cite{nguyen2018automatic}.

\begin{figure}[htp]
    \centering
    \captionsetup{justification=centering}
    \includegraphics[width=\linewidth]{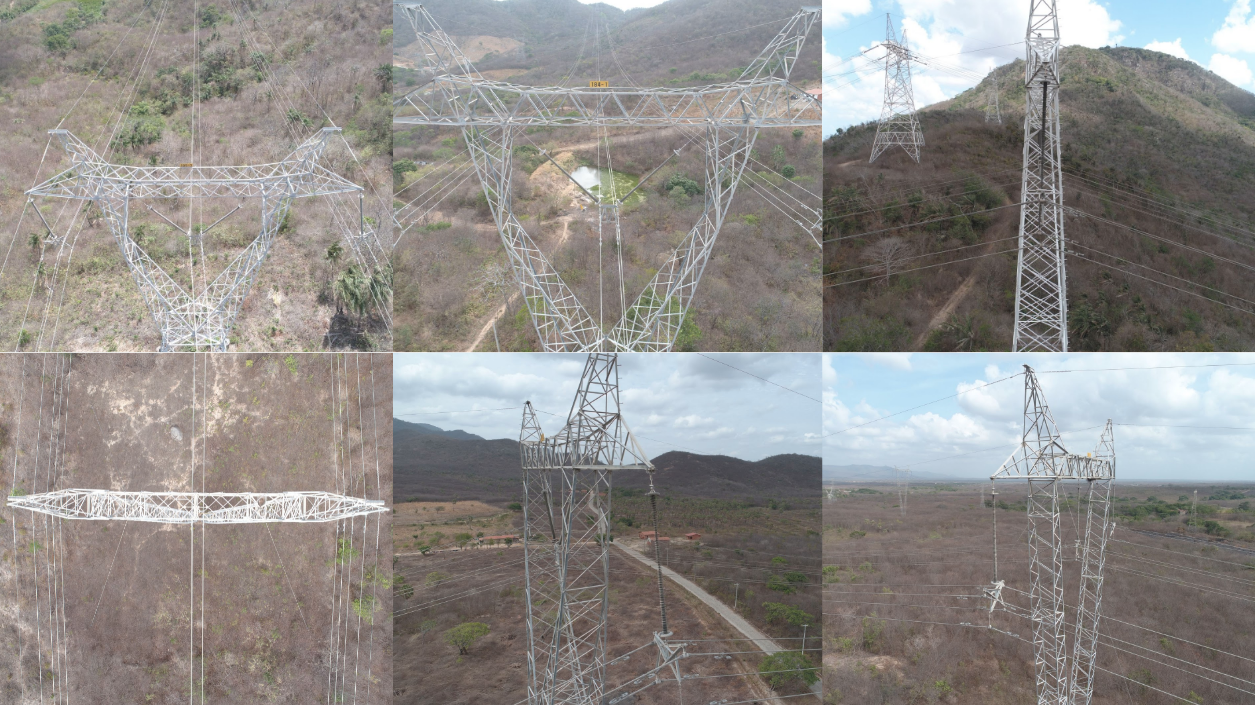}
    \caption{A few image clippings from the proposed dataset.
    }
    \label{fig:dataset}
\end{figure}

Given this scenario, constant maintenance of the equipment is necessary, replacing defective ones before they can cause any loss. Classically, a team has to check each station personally. That implies having someone climbing the transmission tower and checking the condition of its components, which is dangerous and time-consuming \cite{hu2017inspection}. In other words, there is a cost to move people, and it takes time to get up at a station, apart from safety issues associated with the service. For example, Rahmani et al. \cite{rahmani2013descriptive} showed that there were 119 injured workers due to accidents between 2006 and 2012 in an Iranian electricity distribution company, with seven deaths.

Recent developments in computer vision can be applied in the field of Maintenance \& Inspection, improving security and productivity. As an example of application in the security area, 
automatically detecting power line assets via UAVs eliminates much of the safety risks of manual inspection, as inspectors would not need to climb towers as often as before. Instead, much of the inspection could occur through a direct analysis of the detected components, which a human inspector could perform in a secure environment or by a fault classification method \cite{li2021fault}. In addition to the security gains, there are also financial and time gains, as the frequency of moving teams and providing the necessary equipment would decrease.

The dataset plays a major role in the training of deep learning networks, where its quality directly influences the accuracy. That is why we should pay more attention to data, as stated by \cite{sambasivan2021everyone}. Furthermore, public datasets play a fundamental role in a rapidly advancing area. They allow researchers to propose ideas and perform experiments even in scenarios where they cannot get the data by themselves. Also, those datasets usually serve as a benchmark for a specific task, providing a fair comparison among new techniques. 

It is evident how quickly certain areas of computer vision evolved after the introduction of datasets such as CIFAR10 \cite{krizhevsky2009learning}, ImageNet \cite{krizhevsky2012imagenet} and MNIST \cite{lecun1998gradient}. In that sense, public datasets on power line assets for object detection are extremely scarce, and the existing ones are quite limited, typically only supporting one asset type or two, at most \cite{tao2018cplid, tomaszewski2018collection, abdelfattah2020ttpla}. That happens because most of the works in the area are privately funded by companies that want to maintain a competitive advantage by not making their datasets available. 

As the main contribution, this paper introduces a new real-world, high-resolution, and multi-category dataset for multi-size power line assets recognition, the Power Line Assets Dataset, or STN PLAD. It serves as publicly available development data and benchmark for the computer vision community working on automatic power line inspection. 
In addition, experiments with state-of-the-art techniques show the dataset's strengths and limitations. These experiments used two popular general-purpose object detectors, namely SSD and Faster R-CNN. Based on its analysis, a variation of the training pipeline, which we call MS-PAD, is proposed to improve the overall object detection performance in the STN PLAD dataset.



This paper is organized as follows. 
The prior works are presented in \autoref{sec:related}. The properties of the new STN Power Line Assets Dataset are described in \autoref{sec:dataset}. 
\autoref{sec:methodology} shows the methods used to evaluate the proposed dataset.
Next, comparative and performance results of techniques applied in  STN PLAD are presented in \autoref{sec:results}, followed by a discussion about what was seen in the tests in \autoref{sec:discussion} and, lastly, the final remarks in \autoref{sec:conclusion}.

\section{Related works}
\label{sec:related}

This section is divided into two parts. First, the public datasets closely related to power lines are presented, along with their characteristics and limitations. Then, the existing methods that attempt to detect multi-size power line assets in high-resolution images are shown.

\subsection{Public datasets related to power lines}
\label{sec:data-discussion} 

A common problem found in the literature when using Deep Learning to detect power line objects is finding data. There are not enough publicly available datasets to feed detectors based on deep learning methods, or they do not cover enough power line components \cite{nguyen2018automatic, liu2020review}. Many similar works use private datasets, generally provided by the companies or government agencies financing the projects, which tend not to publish them \cite{lei2019intelligent,yang2019insulator,nguyen2018automatic,zhang2019learning,siddiqui2018robust}. Nevertheless, in the literature search presented by Liu et al. \cite{liu2020review} and in the work of Abdelfattah et al. \cite{abdelfattah2020ttpla}, a few publicly available datasets were found but with many limitations. \autoref{tab:datasetreview} summarizes the main public image datasets related to power lines assets for the object detection task. The last line shows the proposed dataset for comparison purposes.

\begin{table*}[htbp]
 \caption{Main public image datasets related to power lines asset detection. 
 }
 \centering
   \begin{threeparttable}
\begin{tabular}{lcccccc}
\toprule
Dataset                                                              & \#Assets                    & Instances/image (average)      & Image size                                                     & Instances & Images & Background variation              \\
\midrule
CPLID \cite{tao2018cplid}                           & 1                           & 1.9                            & 1152$\times$864                                                & 1569                          & 848                        & Limited                           \\
Tomaszewski et al. \cite{tomaszewski2018collection} & 1                           & 1                              & 5616$\times$3744                                               & 2630\tnote{a}                         & 2630\tnote{b}                      & Very limited                      \\
\textbf{STN PLAD}                                      & \textbf{5} & \textbf{18.1} & \textbf{5472$\times$3078 or 5472$\times$3648} & \textbf{2409}                          & \textbf{133}                        & \textbf{Diverse} \\ 
\bottomrule
\end{tabular}
  \begin{tablenotes}
  \item[a] All the instances correspond to the same object.
  \item[b] Images from this dataset are extremely similar and captured from just nine different points of view.
  \end{tablenotes}
  \end{threeparttable}
  \label{tab:datasetreview}
\end{table*}

As evidenced in \autoref{tab:datasetreview}, there is a minimal amount of public datasets related to power line asset detection. They target distinct tasks that can be detection, classification, or segmentation. For instance, the one in \cite{emre2017powerline} is specifically related to conductor wires in low-resolution images for binary classification. Zhang et al. \cite{zhang2019detecting} propose two datasets of binarized masks of conductor wires of power lines in urban and mountain scenarios, respectively. Abdelfattah et al. \cite{abdelfattah2020ttpla} propose a dataset containing pixel-wise annotation (a.k.a. instance segmentation) of both transmission towers and power lines. However, the main competing datasets are CPLID \cite{tao2018cplid}, and the one from Tomaszewski et al. \cite{tomaszewski2018collection} as they are the only ones that use bounding box annotations.

\begin{figure}
\captionsetup{justification=centering}
\centering

\includegraphics[width=\linewidth]{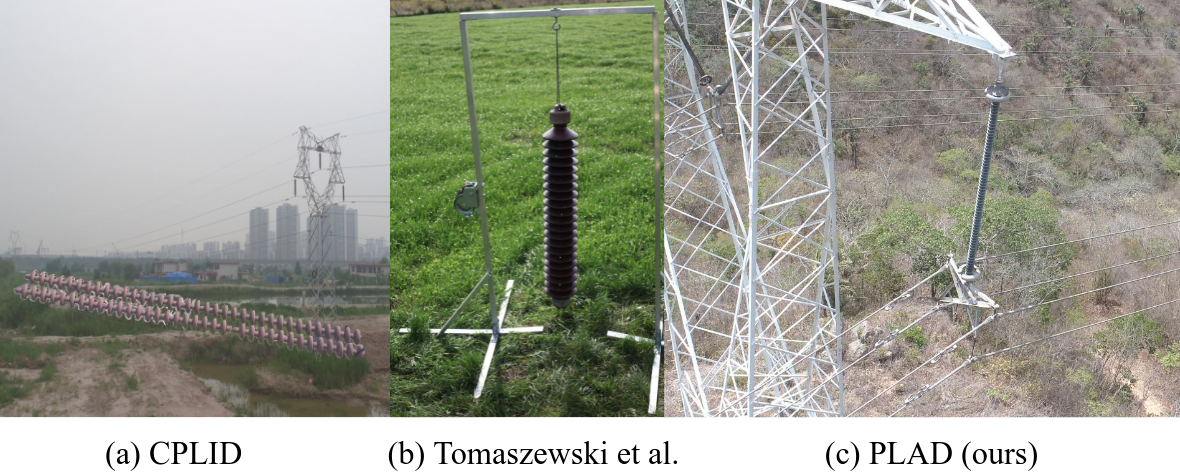}
\caption{Sample images from public datasets as compared to our dataset STN PLAD in which (a) CPLID \cite{tao2018cplid} with an asset inserted in a background, (b) \cite{tomaszewski2018collection} with an asset in a non-real-world scenario, and (c) our dataset (STN PLAD) with multi-objects in real-world scenes (zoomed in image).
}
\label{fig:datasets}
\end{figure}

CPLID \cite{tao2018cplid} is a dataset related only to a specific type of insulator with a specific shape and size. Although it also has annotations of defects in some of those insulators, it lacks a diversity of data since there is only one type of power line asset. The defective samples are also limited because all of them are from data augmentation, i.e., a single faulty insulator was cropped from an image and then pasted into a limited set of backgrounds, like seen in \autoref{fig:datasets}(a). On the other hand, STN PLAD provides asset variability in diverse scenarios.

The dataset in Tomaszewski et al. \cite{tomaszewski2018collection} has even more limitations. They mainly target data quantity (they reported 2630 images) rather than data variability, as they video recorded a ceramic long rod insulator hanging on an apparatus built by them and then extracted some of the frames using a stationary camera. These images only contain one of nine different backgrounds that do not correspond to real-world power line scenarios. An image from one of these nine variations is shown in \autoref{fig:datasets}(b). From the perspective of deep learning techniques, this dataset has a very limited variability of scenes. In summary, although this dataset has a reasonable amount of data, the images taken in the same scenario have a high degree of similarity. Thus, the dataset ends up not being independent and identically distributed (IID) \cite{bottou2010large, rakhlin2012making}, an essential dataset property. In practical terms, it is not feasible to use it in most techniques based on deep learning since it poses little to no challenge to these techniques. In STN PLAD, the images are collected by a drone in the field, providing several real-world scenarios with multiple objects (size, appearance, position, orientation, self-occlusion, background). 

\subsection{Detection of power line assets in high-resolution images}

A few works address the issue of detecting objects related to power lines in high-resolution images. The first one, Zhang et al. \cite{zhang2019learning}, is a study on object detection in high-resolution images captured through Unmanned Aerial Vehicles (UAVs), using deep learning techniques. In their work, the authors propose the MOHR dataset. This private dataset has over ten thousand high-resolution UAV images with five classes: car, truck, building, collapse, and flood damage. The UAV altitudes are high, ranging from 200 to 400 meters, making the objects look quite small. 
The authors apply six general-purpose object detectors, SSD \cite{liu2016ssd} and Faster R-CNN \cite{ren2015faster} included, to the MOHR dataset. The results suggest that detecting small object instances in high-resolution UAV images remains challenging since they perform poorly. The best mean Average Precision (mAP) achieved was 43.94\%, yielded by RFCN-DF \cite{dai2017deformable}. Those results reinforce the relevance of this task.

The works of Kong et al. \cite{kong2018context} and Zhu et al. \cite{zhu2018multi} share some of their authors and contents, indicating that one is an incremental improvement over the other. The former proposes a technique to detect small objects in high-resolution images. 
The technique, based on Faster R-CNN, is tested on a private dataset of 3700 high-resolution images. However, the proposed approach here is limited and prone to issues since only small objects inside the context of a large one are attainable. For instance, dampers are usually small independent objects far from larger ones. Another issue is the low Average Precision (AP) of some classes, such as the tower plate (73.2\%), which the authors justify by saying they are too small.

Finally, Zhu et al. \cite{zhu2018multi} attempts to improve the efficiency of their previous work by merging the two stages in order to share early convolutional layers. 
They also use a private dataset with high-resolution images with six classes: electric tower, vibration damper, spacer, insulator, bird's nest, and tower plate. Despite the efficiency of this method, the same issues and limitations that existed before regarding object detection are maintained. The only difference is that the objects are not so small as their previous work, which is one of the main factors that positively impact the mAP. 

These last two works, \cite{kong2018context} and \cite{zhu2018multi}, are not reproducible since the datasets are private, and they are not open-source.

\section{STN PLAD: Dataset description}


\label{sec:dataset}

The images were captured using a DJI Phantom 4 Pro \footnote{\url{https://www.dji.com/phantom-4-pro}}, and \autoref{fig:dataset} shows some image clippings. A set of policies for data collection was proposed to ensure data variability and consistency. First, the drone was handled by certified drone pilots, who were instructed to capture the images, always maintaining a similar distance to the transmission tower in a wide shot due to the high-resolution nature of the camera. In addition, the drone's viewing angles were varied to ensure better learning by models based on neural networks and diverse daytime, weather, angulation, and illumination conditions. Finally, several transmission towers were captured to obtain background and component variation. This data capture protocol provides images with a wide range and number of power line assets in each one of them, with a mean of 18.1 instances per captured image as can be seen in \autoref{tab:datasetreview} .

The equipped camera is a DJI FC6310 and it can take pictures with a resolution of $5472\times3078$ (3:2) or $5472\times3648$ (16:9). Both aspect ratios were used during data collection.
For annotating the 2409 objects in all 133 captured images, the LabelImg tool was used \cite{tzutalin2015labelimg}. Two annotators were responsible for carefully surrounding each object with a bounding box. Each person took, on average, 10 minutes to annotate one image. Each image is assigned to only one annotator to perform all its annotations. To maintain the annotation consistency between different annotators, they labeled each assigned image with their highest possible scrutiny and were in touch during the entire annotation process.

\begin{table*}[htbp]
 \caption{STN PLAD statistics.}
  \centering
\begin{tabular}{llcccc}
\toprule
Class name              & Label             & Instances & Instances per image & Average Area (px)         & Standard Deviation (px)        \\
\midrule
Transmission tower & \textit{tower}     & 253       & 1.9       & $2.61\times10^6$ & $3.12\times10^6$ \\
Insulator          & \textit{insulator} & 312       & 2.3       & $8.84\times10^4$ & $8.55\times10^4$ \\
Spacer             & \textit{spacer}    & 253       & 1.9       & $2.82\times10^4$ & $2.41\times10^4$ \\
Tower plate        & \textit{plate}     & 86        & 0.6       & $9.42\times10^3$ & $1.11\times10^4$ \\
Stockbridge damper & \textit{damper}    & 1505      & 11.3      & $2.89\times10^3$ & $5.78\times10^3$ \\
    \bottomrule
  \end{tabular}
  \label{tab:objsperclass}
\end{table*}

\begin{figure*}[htp]
\captionsetup{justification=centering}
\centering
\includegraphics[width=0.92\linewidth]{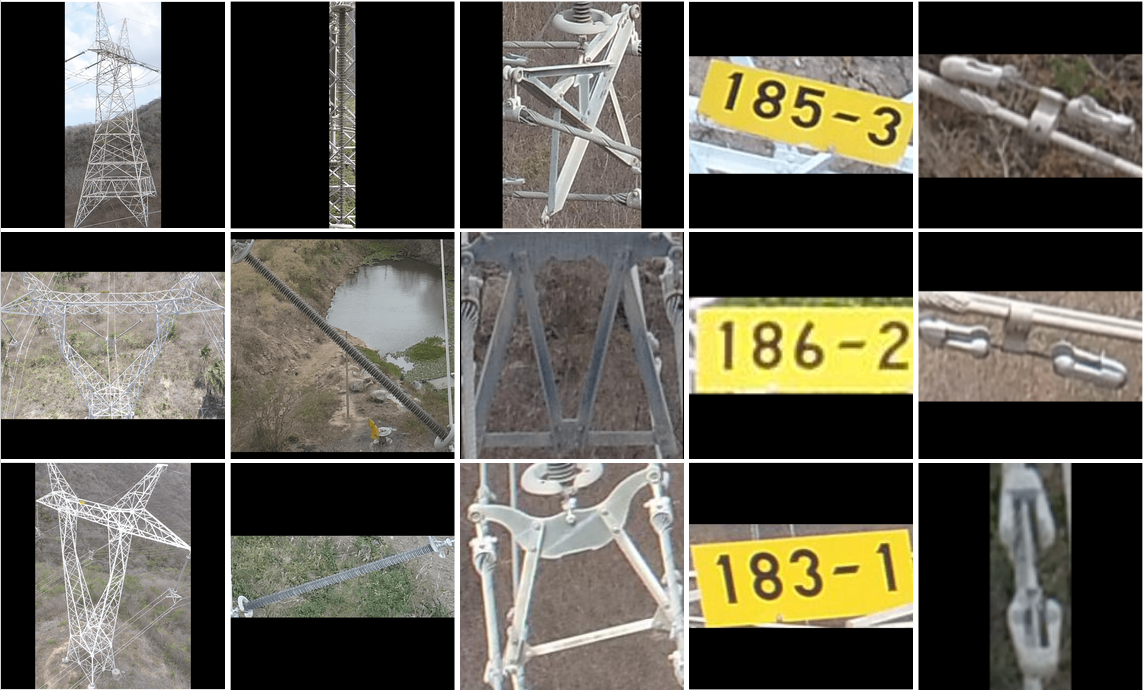}
\caption{Examples of all five classes of power line assets in STN PLAD. Each column shows instances from one class. From left to right: Transmission tower, Insulator, Spacer, Tower plate, and Stockbridge damper.}
\label{fig:assets}
\end{figure*}

The total amount of images in STN PLAD may appear small but, considering the employed data collection protocol, the camera's resolution, and, more importantly, the total amount of object instances, it can be seen that it has a reasonable amount of data. Images from STN PLAD have considerably more information than regular images from common datasets, such as ImageNet \cite{krizhevsky2012imagenet} and MSCOCO \cite{lin2014microsoft}. On average, the STN PLAD has more than 18 objects per image with an average area of at least $2.89\times10^3$ pixels. This 18 objects/image density is way above the related datasets, as seen in \autoref{tab:datasetreview}. Finally, the STN Power Line Assets Dataset is publicly available in \url{https://github.com/andreluizbvs/PLAD} \footnote{In case the article is accepted, the dataset will be posted on a web page with a structured presentation.}.

\section{Methods}
\label{sec:methodology}

This section describes two techniques that are often used to validate object detection datasets \cite{tao2018cplid}, SSD and Faster R-CNN. Their performance on STN PLAD is presented in the next section and discussed later. The observed limitations in dealing with the proposed dataset inspired the creation of a pipeline called MS-PAD, which is also detailed here.

\subsection{Single Shot MultiBox Detector (SSD)}
\label{sec:ssd}

In the context of power line inspection, SSD is one of the suggested techniques of two recent reviews \cite{liu2020review, nguyen2018automatic} to target the problem of detecting assets on power transmission towers. Both reviews have the same context as this work, focusing on inspecting power line assets from UAV images. Moreover, they analyze Deep Learning techniques applied to solve problems in the area. Some of the mentioned problems are assets detection, assets segmentation, assets fault identification.

The parameters of the SSD used are the same as in the original work by Liu et al. \cite{liu2016ssd}, such as the backbone, VGG16, and all the parameters and dimensions for the convolutional layers. In the original work, two different input layers were proposed, $300\times300$ and $512\times512$. In our experiment, the latter was used since it achieved better accuracy than the former, according to the original results. Also, the images used for this experiment have a higher resolution. This high-resolution implies that the larger the size of the input layer, the less the resizing effect will affect the input image quality. 
Finally, weights pre-trained with the COCO Dataset \cite{lin2014microsoft} were used. 


\subsection{Faster R-CNN}
\label{sec:fasterrcnn}
The objective of including the Faster R-CNN \cite{ren2015faster} in the tests was to use a recent technique of object detection to obtain results close to the current state of the art. 
Faster R-CNN-based networks are also suggested to detect and inspect power transmission towers according to the same reviews mentioned in \autoref{sec:ssd}. They are also well consolidated, have performed well in object detection competitions, and are used by similar works \cite{liu2020review, nguyen2018automatic}. 

The network used for this experiment was the Feature Pyramid Network (FPN) Faster R-CNN \cite{lin2017feature}. FPN aims to improve the detection of small objects, as it uses multi-scale feature maps and higher resolution layers to build new semantically rich layers. Thus, information from the initial layers is used. These layers are traditionally less condensed, but even so, they already have a high semantic level. ResNet-101 was used as a backbone, which had the best detection result in its publication \cite{lin2017feature}. Also, the input resolution was chosen in order to decrease the image resizing impact. The input image is resized to $2736\times1824$, representing a downscaling factor of $4$ when compared to the original size, which is much less than the downscaling factor of approximately $19$ used in the SSD experiment. All other parameters were kept as the original. 
    
    


\subsection{MS-PAD}
\label{sec:ourmethod}

\begin{figure}[htp]
    \centering
    \includegraphics[width=0.95\linewidth]{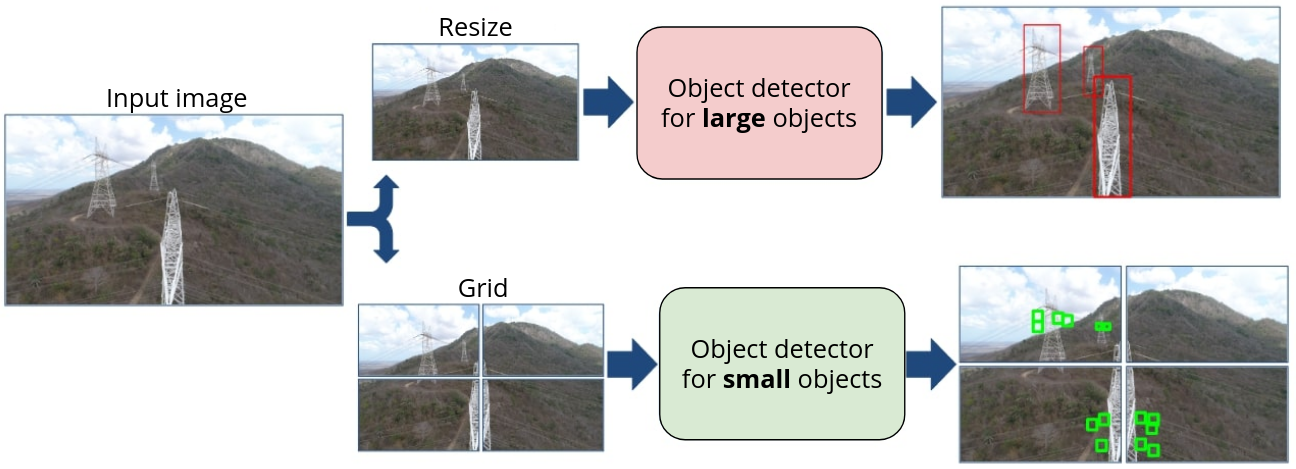}
    \caption{MS-PAD pipeline. The input image is resized for the first network. Meanwhile, the image is divided into a grid, resulting in the second network's inputs that generate different bounding boxes.}
    \label{fig:method}
\end{figure}

After observing the results related to the SSD and Faster R-CNN methods, it was noticed that a simple pipeline modification could enhance the overall performance of power line assets detection in STN PLAD. This approach takes advantage of the images' high resolution, where information is lost after resizing. In the Multi-Size Power line Asset Detection (MS-PAD) workflow, represented in \autoref{fig:method}, two independent networks are trained separately. The SSD was chosen because it performed better, as it will be shown in the next section.

The first of these two networks uses the original pipeline that resizes its input, but this time trained without the Stockbridge damper class. The damper asset was excluded because it has a much smaller size, being harder to identify after image resizing. This strategy can be applied to other assets not contemplated in this work that are small. An important note is that, although the tower plate is also small, it still had enough features after resizing that made it highly recognizable.

The second network is responsible for detecting small objects, in this case, the Stockbridge damper class. It goes through a different process, where the initial image is split following a grid. The image is divided into 16 smaller ones with a fixed resolution of $1368\times769$ or $1368\times912$, depending on the original image, which can be $5472\times3078$ (3:2) or $5472\times3648$ (16:9). This $4\times4$ division is constant since the drone pilots followed a data collection protocol, in which the drone had to stay at similar distances from the transmission towers, as described in \autoref{sec:dataset}. 

In summary, only the Stockbridge damper class is submitted to the second step of MS-PAD, which divides the high-resolution image in a $4\times4$ grid. This choice is based on the average area of the classes of \autoref{tab:objsperclass} and the AP in the original image-resizing approach
. The Stockbridge damper has the lowest average area and was not well detected using the previously mentioned methods, indicating the need to receive extra attention compared to other classes.



\section{Experimental results}
\label{sec:results}

This section presents the two conducted experiments and their results. The first one is responsible for comparing the performance of the two mentioned object detectors in the proposed dataset. The second one demonstrates another way to deal with the input data, using MS-PAD, which was detailed in \autoref{sec:ourmethod}. In all experiments, the standard metric of Average Precision (AP) is used to evaluate object detection performance on STN PLAD. 
In order to validate and obtain a greater degree of confidence in the results of the proposed MS-PAD, the Monte Carlo cross-validation method \cite{dubitzky2007fundamentals} was chosen and implemented in its experiments presented in \autoref{sec:mspad-comparison}. This method creates $k$ random splits of train and test sets of the whole dataset. Then, the model is trained and tested for each $k$ split, and the final result is the average. In the end, this section shows which object detector and which pipeline present the best results in the described scenario according to the considered metric.

For the experiments, STN PLAD was split in a standard 80/20 proportion for the training and test sets, respectively. Also, to consider that an object was correctly detected, the Intersection over Union (IoU) between the ground truth and the predicted bounding box had to be equal to or larger than 0.5. Also, data augmentation is already an embedded stage in both implementations of the techniques. Finally, the experiments were performed on a desktop running the Ubuntu 18.04 Operating System, powered by an Nvidia RTX 2080 Ti GPU (11 GB of VRAM) and an Intel Core i7 - 4790K CPU @ 4.00 GHz with 32 GB of available RAM. 


\subsection{SSD and Faster R-CNN results}
\label{sec:comparison}


For this test, both detectors were trained once and for the same period, about two days. The mAP results of using the MS-PAD approach for SSD and Faster R-CNN were 90.2\% and 88.6\%, respectively, showing that SSD has a slight advantage over Faster R-CNN. 
These methods were also applied to the two main dataset competitors of the proposed STN PLAD. In CPLID \cite{tao2018cplid}, SSD and Faster R-CNN achieved 98.17\% and 98.31\% mAP, respectively. Regarding Tomaszewski et al. \cite{tomaszewski2018collection}, both detectors reached 100\% mAP.

\begin{figure}
    \centering
    \includegraphics[width=0.9\linewidth]{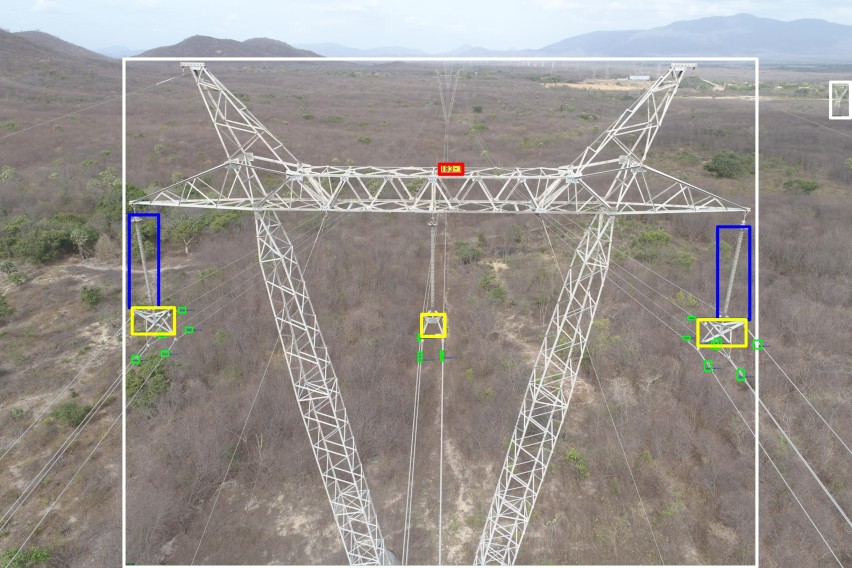}
    \caption{Qualitative detection results of the SSD technique using the original image-resizing approach.}
    \label{fig:ssd-resize}
\end{figure}

\begin{figure}
    \centering
    \includegraphics[width=0.9\linewidth]{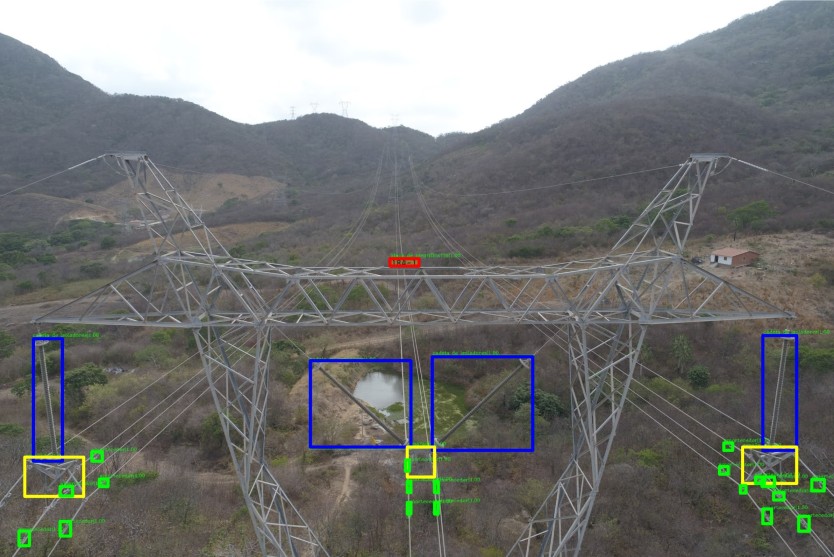}
    \caption{Qualitative detection results of the Faster R-CNN technique using the original image-resizing approach.}
    \label{fig:frcnn-resize}
\end{figure}

\autoref{fig:ssd-resize} and \autoref{fig:frcnn-resize} show the visual results regarding the detected objects by the SSD and the Faster R-CNN methods, respectively, using the original approach, which only resizes the input image. In the images, the bounding boxes' colors are connected with the dataset classes: blue is for the Insulators; yellow is for the Spacers; green is for the Stockbridge dampers; red is for the Tower plate; white is for the Transmission tower. It is possible to see in both figures how small the Stockbridge dampers are related to other objects. These images also illustrates failure cases, like the middle insulator in \autoref{fig:ssd-resize} and the transmission tower in \autoref{fig:frcnn-resize}.

\subsection{MS-PAD results}
\label{sec:mspad-comparison}



For this experiment, $k = 5$, so five splits with randomly selected samples were used, in which each split is used twice: one time for the original image-resizing approach and another time for MS-PAD. The total amount of iterations for each training session is fixed at 20,000. The obtained results regarding AP for each split are shown in \autoref{tab:montecarlo-map}. In addition, \autoref{tab:resizevsproposed} shows the average results from \autoref{tab:montecarlo-map} reached by each approach side-by-side, in a direct comparison.

\begin{table*}[htpb]
\centering
\caption{Comparison of detection results of the original image-resizing approach (Original) and the MS-PAD pipeline (Ours) of each Monte Carlo cross-validation split ($k$) relative to the Average Precision (AP). The best mAP results are in bold.}
\begin{tabular}{lcccccccccc}
\toprule
  & \multicolumn{2}{c}{$k = 1$}                       & \multicolumn{2}{c}{$k = 2$}                       & \multicolumn{2}{c}{$k = 3$}                       & \multicolumn{2}{c}{$k = 4$}                       & \multicolumn{2}{c}{$k = 5$}                       \\
  \cmidrule(lr){2-3} \cmidrule(lr){4-5} \cmidrule(lr){6-7} \cmidrule(lr){8-9} \cmidrule(lr){10-11}
                  & \multicolumn{1}{l}{Original} & \multicolumn{1}{l}{Ours} & \multicolumn{1}{l}{Original} & \multicolumn{1}{l}{Ours} & \multicolumn{1}{l}{Original} & \multicolumn{1}{l}{Ours} & \multicolumn{1}{l}{Original} & \multicolumn{1}{l}{Ours} & \multicolumn{1}{l}{Original} & \multicolumn{1}{l}{Ours} \\
\midrule
Transmission tower & 0.885                      & 0.905                    & 0.883                      & 0.901                    & 0.875                      & 0.874                    & 0.920                      & 0.883                    & 0.945                      & 0.938                    \\
Insulator          & 0.825                      & 0.938                    & 0.924                      & 0.866                    & 0.931                      & 0.893                    & 0.839                      & 0.884                    & 0.874                      & 0.889                    \\
Spacer             & 0.917                      & 0.810                    & 0.789                      & 0.850                    & 0.914                      & 0.910                    & 0.863                      & 0.805                    & 0.853                      & 0.905                    \\
Tower plate        & 0.932                      & 0.994                    & 0.830                      & 1.00                     & 0.941                      & 0.990                    & 0.984                      & 0.997                    & 0.995                      & 0.876                    \\
Stockbridge damper & 0.189                      & 0.829                    & 0.201                      & 0.882                    & 0.189                      & 0.870                    & 0.264                      & 0.824                    & 0.227                      & 0.787                    \\
\midrule
mAP                & 0.750                      & \textbf{0.895}                    & 0.725                      & \textbf{0.900}                    & 0.770                      & \textbf{0.907}                    & 0.774                      & \textbf{0.879}                    & 0.779                      & \textbf{0.879}           
         \\
\bottomrule        
\end{tabular}
\label{tab:montecarlo-map}
\end{table*}

\begin{table}[htpb]
\centering
\caption{Detection average results from \autoref{tab:montecarlo-map} of both approaches, side-by-side. The best results for each class and the mAP is in bold.}
\begin{tabular}{lcc}
\toprule
                   & \multicolumn{1}{c}{Original} & \multicolumn{1}{c}{Ours} \\
\midrule
Transmission tower & \textbf{0.902}                      & 0.900                    \\
Insulator          & 0.879                      & \textbf{0.894}           \\
Spacer             & \textbf{0.867}                      & 0.856                    \\
Tower plate        & 0.936                      & \textbf{0.971}           \\
Stockbridge damper & 0.214                      & \textbf{0.838}           \\
\midrule
mAP                & 0.755                      & \textbf{0.892}          \\
\bottomrule
\end{tabular}
\label{tab:resizevsproposed}
\end{table}

\begin{figure}
    \centering
    \includegraphics[width=0.9\linewidth]{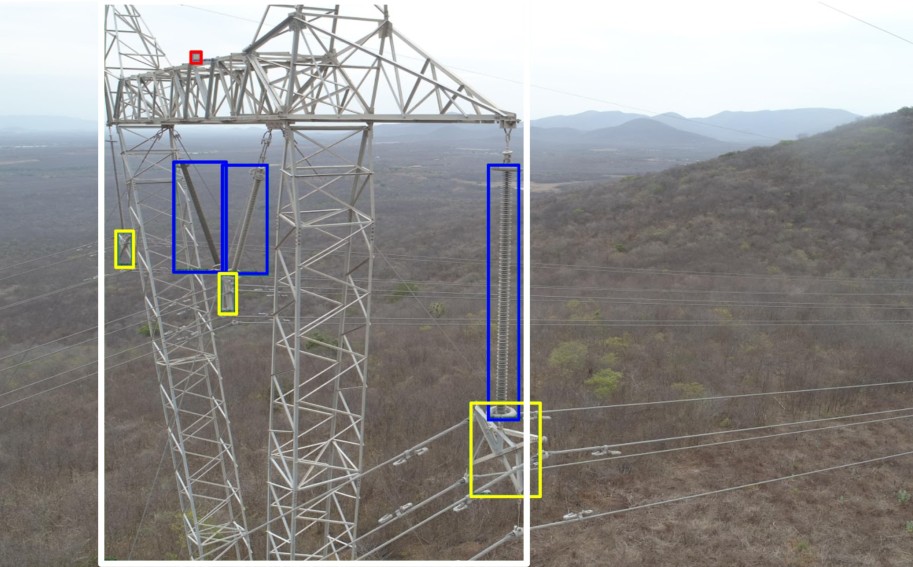}
    \caption{Qualitative detection results of the MS-PAD pipeline for the big object classes.}
    \label{fig:twonetworks-big}
\end{figure}

\begin{figure}
    \centering
    \includegraphics[width=0.9\linewidth]{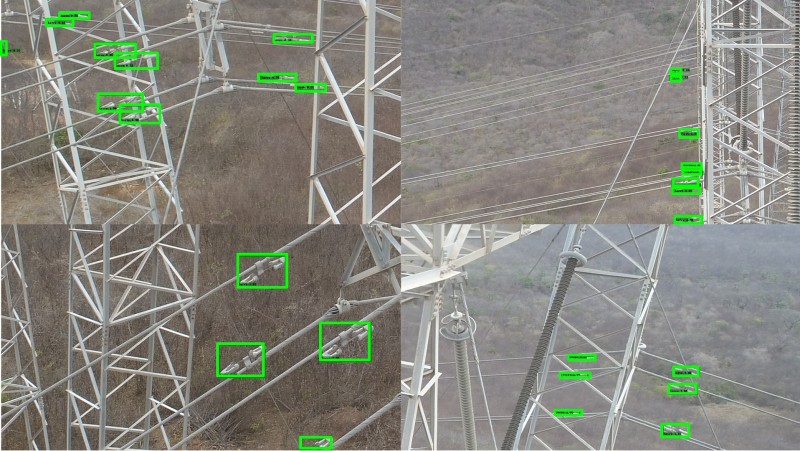}
    \caption{Qualitative detection results of the MS-PAD pipeline for the small object classes.}
    \label{fig:twonetworks-small}
\end{figure}

\autoref{fig:twonetworks-big} and \autoref{fig:twonetworks-small} present the qualitative performance of MS-PAD for big and small objects, respectively. The color codes previously mentioned are maintained for these images.

\section{Discussion}
\label{sec:discussion}

This section details and discusses all results presented in \autoref{sec:results}, also giving insights into the usage of the MS-PAD approach in the proposed STN Power Line Asset Dataset.

\subsection{STN PLAD strengths \& limitations}

The proposed STN PLAD is the first public power line assets dataset with multiple objects in real-world scenarios. It contains five classes of entirely different objects with multiple instances each, in varied real backgrounds. The data collection protocol allows for a balance in data quantity and variability since the captured images vary in illumination, backgrounds, and weather conditions. Also, the drone position is not fixed in order to obtain objects data from different perspectives. Another STN PLAD challenging characteristic is that there are many objects per image (18.1, on average) compared to the related public datasets, which commonly have a small instance per image rate (1 and 1.9, on average). Thanks to this process, STN PLAD poses a reasonable challenge to recent deep learning techniques, as observed in the \autoref{sec:results}, in which the best of the tested approaches achieved an 89.2\% mAP, leaving considerable room for improvement.

Although the proposed STN PLAD provides new grounds in the power line area and stimulates the development of power line asset detection methods, it still has limitations. The main one is related to its total amount of images. That prevents some data-hungry object detectors from performing successfully since they would require a more extensive dataset. Another disadvantage is that the images were only collected from one private transmission line. Even though different transmission lines tend to be similar, it would be better to have images of several transmission lines in other places to reduce the bias of background, environment, and electrical assets appearance. Finally, the images belong to a power line in Brazil, which may not apply to other countries.

\subsection{SSD and Faster R-CNN comparison in STN PLAD}

This discussion is related to the experiment in \autoref{sec:comparison}. 
According to the proposed methodology, when performing this experiment, it was expected that the results related to the Faster R-CNN would surpass the results from the SSD network considering the applied metrics. However, it can be observed in the results reported in \autoref{sec:comparison} 
that it did not occur. This result was obtained due to the limitations of the used data. Deep learning techniques benefit from the use of large amounts of data. According to Ng \cite{ng2017machine} \cite{tang2018canadian} when the number of data limits a deep learning technique, shallower techniques can obtain comparable or even better results than deeper techniques. The used Faster R-CNN is much deeper and has more trainable parameters than the used SSD network. Therefore, for a limited amount of data, the learning of the used Faster R-CNN is limited.

The other results in this comparison were related to the SSD and Faster R-CNN performance in the competing datasets. In \cite{tomaszewski2018collection}, both methods achieved 100\% mAP, as expected due to the reasons presented in \autoref{sec:data-discussion}. In \cite{tao2018cplid}, the original SSD and Faster R-CNN obtained performances above 98\% mAP. The high mAP values obtained by both object detectors in both competing datasets showed how well-resolved their challenges already are.

\subsection{MS-PAD in STN PLAD}

It is possible to see in \autoref{tab:montecarlo-map} the mAP values reached by MS-PAD are higher than all mAP values of the Original approach in at least ten percentage points ($k=5$) and at most 17.5 percentage points ($k=2$).
\autoref{tab:resizevsproposed} shows a direct comparison, in which the values for each approach are an average of the five splits showed in \autoref{tab:montecarlo-map}. The best values for each class AP and mAP are in bold. MS-PAD yields the best AP result in three out of the five total classes, and there is a gap of 13.7 percentage points between mAPs. That gap is primarily due to the Stockbridge damper AP improvement, which grew 62.4 percentage points using MS-PAD. 

It is noteworthy that the performance of large objects changes when comparing the Original and the MS-PAD. That may happen because during the MS-PAD resize branch training, one less class is considered (Stockbridge damper). That directly influences how the network learns since there is a different amount of objects and classes, directly impacting the final performance of large assets. Also, it is important to note that there is no guarantee that the performance impact will be positive or negative when training with one less class.

\section{Conclusions}
\label{sec:conclusion}

This work proposes a new public real-world high-resolution power line asset dataset with multiple assets categories, called STN Power Line Assets Dataset (PLAD). Its images were captured by an Unmanned Aerial Vehicle (UAV) following a data collection protocol to ensure data variability in order to benefit deep learning models. STN PLAD contains 2409 annotated objects across 133 images divided into five classes with different shapes and sizes. It has the biggest amount of power line asset types among all public power line assets datasets, with the highest density of objects per image between them as well. The latter is possible due to its images having far above average resolutions, 5472$\times$3078 and 5472$\times$3648, more precisely. After evaluating STN PLAD in recent general-purpose object detectors, a different pipeline called MS-PAD is proposed. This pipeline contains a simple modification that allows for an mAP improvement from 75.5\% to 89.2\%. STN PLAD is publicly available to mitigate the lack of data in the power line inspection area and provide a new challenge to the computer vision community in order to stimulate the proposition of new asset detection methods for power lines.

\section*{Acknowledgment}
The authors acknowledge the financial support of STN - Sistema de Transmissão Nordeste S.A. through the ANEEL R\&D Program for the development of development of the research project entitled: “PD-04825-0006/2019: Inspeção com Drones por Meio do Acoplamento Eletrostático para Carregamento de Baterias em Voo e Uso de Aprendizagem de Máquina para Classificação Automática de Defeitos”. 

This research was funded in part by the Coordenação de Aperfeiçoamento de Pessoal de Nível Superior - Brasil (CAPES) - Finance Code 001 and by the Conselho Nacional de Desenvolvimento Científico e Tecnológico (CNPq).




\bibliographystyle{IEEEtran}
\bibliography{main}
%
%


\end{document}